%% file: main.tex
\documentclass[10pt,twocolumn,letterpaper]{article}

\usepackage{iccv}              


\definecolor{iccvblue}{rgb}{0.21,0.49,0.74}
\usepackage[pagebackref,breaklinks,colorlinks,allcolors=iccvblue]{hyperref}
\usepackage[accsupp]{axessibility}

\usepackage{graphicx}
\graphicspath{{/images/}}
\usepackage{amsmath}
\usepackage{amssymb}
\usepackage{booktabs}
\usepackage{multicol}
\usepackage{multirow}
\usepackage{algorithm}
\usepackage{algpseudocode}
\usepackage{comment}
\usepackage{colortbl}
\usepackage{array}
\usepackage[inkscape]{svg}
\usepackage{placeins}
\definecolor{LightYellow}{RGB}{255, 249, 196}
\definecolor{LightGray}{gray}{0.92}
\definecolor{RoyalBlue}{RGB}{65,105,225}
\definecolor{Emerald}{RGB}{80,200,120}
\definecolor{DeepPink}{RGB}{255,20,147}
\definecolor{Yellow}{RGB}{255,255,0}
\definecolor{customYellow}{HTML}{FFD966}
\definecolor{customGreen}{rgb}{0.88, 1, 0.88} 
\usepackage{orcidlink}

\usepackage{subcaption}  
\usepackage{tablefootnote}

\def\paperID{84} 
\def\confName{ICCV}
\def\confYear{2025}

\newcommand{\methname}{DiffAug\xspace}

\title{Diffusion-Based Data Augmentation for Medical Image Segmentation}

\author{Maham Nazir$^{1,\dagger}$ \orcidlink{0009-0004-1832-297X} \and Muhammad Aqeel$^{2,\dagger}$ \orcidlink{0009-0000-5095-605X} \and Francesco Setti$^2$ \orcidlink{0000-0002-0015-5534} \\
$^1$School of Computer Science and Engineering, Beihang University\\
Beijing, China\\
$^2$Dept. of Engineering for Innovation Medicine, University of Verona\\
Strada le Grazie 15, Verona, Italy\\
{\tt\small Contact author: muhammad.aqeel@univr.it}
}

\begin{document}
\maketitle
\let\thefootnote\relax\footnotetext{$^\dagger$Equal contribution.}
\input{sec/0_abstract}    
\input{sec/1_intro}

\input{sec/2_formatting}
\input{sec/3_finalcopy}
{
    \small
    \bibliographystyle{ieeenat_fullname}
    \bibliography{main}
}

\end{document}

%% file: sec/0_abstract.tex
\begin{abstract}
Medical image segmentation models struggle with rare abnormalities due to scarce annotated pathological data. We propose \methname a novel framework that combines text-guided diffusion-based generation with automatic segmentation validation to address this challenge. Our proposed approach uses latent diffusion models conditioned on medical text descriptions and spatial masks to synthesize abnormalities via inpainting on normal images. Generated samples undergo dynamic quality validation through a latent-space segmentation network that ensures accurate localization while enabling single-step inference. The text prompts, derived from medical literature, guide the generation of diverse abnormality types without requiring manual annotation. Our validation mechanism filters synthetic samples based on spatial accuracy, maintaining quality while operating efficiently through direct latent estimation. Evaluated on three medical imaging benchmarks (CVC-ClinicDB, Kvasir-SEG, REFUGE2), our framework achieves state-of-the-art performance with 8-10\% Dice improvements over baselines and reduces false negative rates by up to 28\% for challenging cases like small polyps and flat lesions critical for early detection in screening applications.

\end{abstract}

\noindent \textbf{Keywords:} Text-guided generation, Spatial conditioning, Segmentation networks, Synthetic data

%% file: sec/1_intro.tex
\section{Introduction}
\label{sec:intro}

Medical image analysis has emerged as a cornerstone of modern healthcare, enabling early disease detection, precise diagnosis, and personalized treatment planning~\cite{litjens2017survey}. Deep learning models have demonstrated remarkable success in automating these tasks, achieving expert-level performance in various clinical applications~\cite{esteva2021deep}. However, the effectiveness of these models critically depends on the availability of large-scale, well-annotated datasets that adequately represent the full spectrum of anatomical variations and clinical abnormalities~\cite{willemink2020preparing}. This dependency poses a fundamental challenge: while normal cases are abundant in clinical databases, pathological findings—such as polyps in colonoscopy or glaucomatous changes in fundus images—remain scarce due to their low prevalence, privacy concerns, and the substantial cost of expert annotations~\cite{campanella2019clinical}.

The scarcity of abnormal training data creates severe class imbalance that significantly impacts model performance, leading to reduced sensitivity for rare pathologies where early detection is crucial~\cite{johnson2019survey}. Traditional data augmentation techniques, such as rotation, scaling, and elastic deformations, fail to address this imbalance as they cannot synthesize novel pathological patterns~\cite{shorten2019survey}. While generative adversarial networks (GANs) have been explored for medical image synthesis~\cite{yi2019generative, kazeminia2020gans}, they suffer from mode collapse, training instability, and limited control over the spatial location and appearance of generated abnormalities~\cite{arjovsky2017wasserstein}. Consequently, existing approaches cannot reliably generate diverse, anatomically plausible abnormalities with precise spatial control, limiting their utility for medical image augmentation.

The advent of diffusion models has revolutionized generative modeling, demonstrating unprecedented quality in image synthesis tasks~\cite{dhariwal2021diffusion}. These models shown remarkable ability to generate high-fidelity images with fine-grained control through multimodal conditioning~\cite{rombach2022high}. In the medical domain, initial explorations of diffusion models have focused primarily on image-to-image translation and reconstruction tasks~\cite{kazerouni2023diffusion, wolleb2022diffusion}. However, their potential for controlled anomaly synthesis—leveraging both spatial masks and textual descriptions—remains largely unexplored. Furthermore, existing approaches lack robust validation mechanisms to ensure that generated anomalies are not only visually realistic but also accurately localized~\cite{muller2023multimodal}.

To address these limitations, We present a novel framework \methname (\emph{Diffusion-Based Data Augmentation}) that bridges this critical gap by combining text-guided diffusion-based generation with automatic segmentation validation for medical data augmentation. Our approach leverages latent diffusion models to synthesize abnormalities through a controlled process: we utilize descriptive text prompts based on medical literature and anatomical descriptions (e.g., "polyp with irregular surface" or "enlarged optic cup") combined with spatial masks derived from existing annotations. The generated abnormalities are validated using an efficient segmentation network that operates with single-step inference, ensuring spatial consistency. This validation mechanism filters generated samples based on localization accuracy, maintaining quality while enabling efficient processing.

The key contributions of our paper are summarized as follows:
\begin{itemize}
\item A text-guided synthesis framework that leverages anatomical descriptions and spatial conditioning to generate controlled abnormalities for medical image augmentation.
\item An integrated pipeline combining diffusion-based generation with dynamic segmentation validation, ensuring generated abnormalities are properly localized within specified regions.
\item Experimental validation on CVC-ClinicDB, Kvasir-SEG, and REFUGE2 datasets showing improved segmentation performance when training with synthetic data augmentation.
\item An efficient implementation that balances generation quality with computational practicality, making the approach viable for clinical research settings.
\end{itemize}


\section{Related Work}
\label{sec:relatedwork}

Traditional augmentation techniques in medical imaging rely on geometric transformations that cannot address class imbalance by synthesizing new abnormal cases~\cite{shorten2019survey}. Generative Adversarial Networks (GANs) have been extensively explored for medical image synthesis~\cite{yi2019generative}, with variants like StyleGAN achieving high-resolution generation~\cite{karras2019style}. However, GANs suffer from mode collapse and training instability, limiting their ability to generate diverse, controllable abnormalities~\cite{kazeminia2020gans}. VAEs and normalizing flows offer more stable training but produce lower-quality images with limited detail preservation~\cite{kingma2013auto, pawlowski2020deep}.
Recent diffusion models have revolutionized generative modeling, surpassing GANs in both quality and stability~\cite{dhariwal2021diffusion}. In medical imaging, diffusion models have primarily been applied to reconstruction tasks~\cite{jalal2021robust} and image-to-image translation~\cite{kazerouni2023diffusion}. While Wolleb et al. explored diffusion for segmentation ensembles~\cite{wolleb2022diffusion}, and MedSegDiff proposed dynamic conditioning~\cite{wu2024medsegdiff}, these focus on segmentation rather than generation. Notably, text-guided medical generation remains underexplored—RoentGen demonstrated text-to-X-ray synthesis but lacked spatial control~\cite{chambon2022roentgen}. Our work addresses this gap by introducing text prompts with spatial conditioning for controlled abnormality synthesis.

Anomaly detection methods typically use reconstruction-based approaches, identifying abnormalities through high reconstruction errors on models trained with normal data~\cite{10.1007/978-3-319-59050-9_12, baur2021autoencoders}. Despite advances in self-supervised and contrastive learning~\cite{shiri2025madclip, schluter2022natural, aqeel2025CoMet}, these methods remain fundamentally limited by scarce abnormal training examples.
For polyp segmentation, U-Net variants with attention mechanisms dominate~\cite{fan2020pranet}, while transformer architectures show promise for global context modeling~\cite{wang2022stepwise}. In glaucoma assessment, multi-task learning simultaneously segments optic structures and predicts disease risk~\cite{fu2018joint, orlando2020refuge}. However, both domains struggle with underrepresented cases—small polyps and subtle glaucomatous changes—due to limited training data~\cite{silva2014toward}.

While natural language integration has advanced medical image understanding through Visual Question Answering~\cite{zhan2020medical} and CLIP-based zero-shot classification~\cite{zhang2023large}, text-guided medical image generation remains nascent. Unlike natural image domains where DALL-E 2 and Stable Diffusion enable rich text-controlled generation~\cite{rombach2022high, aqeel2025latent}, medical applications lack frameworks that leverage textual descriptions for synthesis. Our work bridges this gap by incorporating text-based guidance for abnormality generation, combined with automatic validation to ensure spatial accuracy—creating a practical solution for text-guided medical data augmentation.

\section{\methname Pipeline}
\label{sec:method}

\begin{figure*}[t]
\centering
\includegraphics[width=\linewidth]{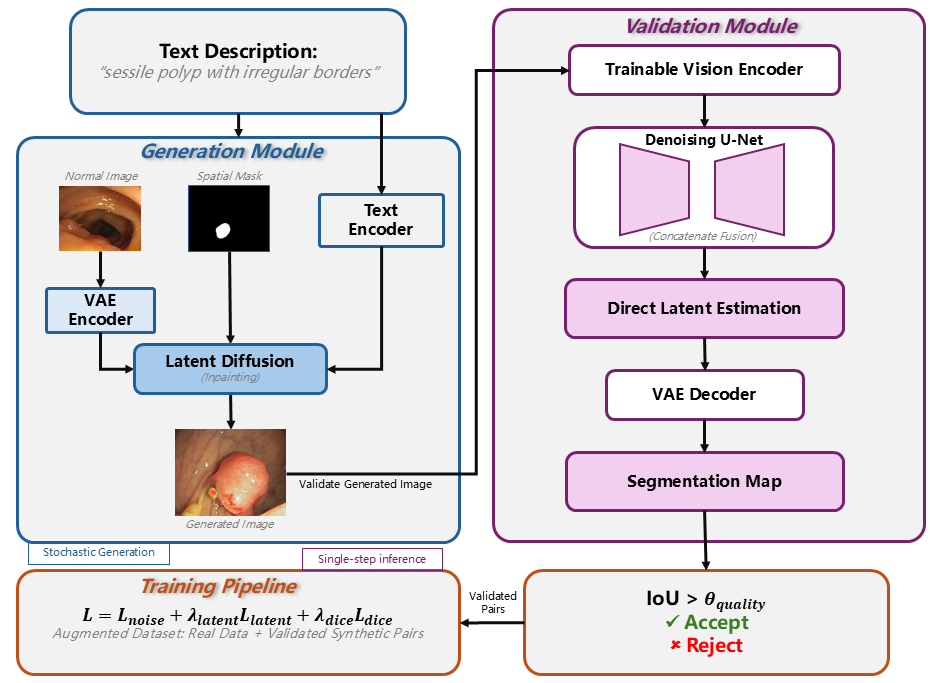}
\caption{Overview of our integrated generation-validation pipeline. \textbf{Left:} Text-guided data generation module uses text prompts and spatial masks to generate abnormalities via latent diffusion inpainting on normal images. \textbf{Right:} Latent-space segmentation validation ensures generated abnormalities are accurately localized through single-step inference. The quality gate accepts only samples with $\text{IoU} > \theta_{\text{quality}}$ for training augmentation.}
\label{fig:pipeline}
\end{figure*}

Our framework addresses the critical challenge of abnormal data scarcity in medical imaging through an integrated pipeline that combines language-guided diffusion-based generation with segmentation validation, as shown in Figure~\ref{fig:pipeline}. We demonstrate that textual descriptions derived from medical terminology can guide abnormality generation when combined with spatial masks, while maintaining quality, a self-validating mechanism that ensures both visual realism and anatomical accuracy.

\subsection{Language-Guided Diffusion Based Image Generation}
We leverage the power of latent diffusion models (LDMs) to generate controlled medical abnormalities, operating in a compressed latent space that balances computational efficiency with preservation of fine-grained medical details. As shown in Figure~\ref{fig:pipeline}, our method performs targeted synthesis by inpainting abnormalities onto healthy tissue, ensuring anatomical consistency while introducing variations.

Given a dataset of normal medical images $I_n = \{I_n^{(i)}\}_{i=1}^N$ where $I_n \in \mathbb{R}^{H \times W \times 3}$, we first encode each image into a lower-dimensional latent representation using a pre-trained variational autoencoder (the VAE from Stable Diffusion). The encoder $\mathcal{E}: \mathbb{R}^{H \times W \times 3} \to \mathbb{R}^{h \times w \times d}$ compresses the image by a factor of $f = \frac{H}{h} = \frac{W}{w}$, yielding:
\begin{equation}
z_n = \mathcal{E}(I_n)
\end{equation}

The core of our generation process employs denoising diffusion probabilistic models adapted for conditional inpainting. The forward diffusion process progressively corrupts the latent representation through a Markov chain:
\begin{equation}
q(z_t | z_{t-1}) = \mathcal{N}(z_t; \sqrt{1-\beta_t} z_{t-1}, \beta_t \mathbf{I})
\end{equation}

Through the reparameterization trick, we can directly sample any intermediate state:
\begin{equation}
z_t = \sqrt{\bar{\alpha}_t} z_0 + \sqrt{1 - \bar{\alpha}_t} \epsilon, \quad \epsilon \sim \mathcal{N}(0, \mathbf{I})
\end{equation}
where $\bar{\alpha}_t = \prod_{s=1}^t (1 - \beta_s)$ represents the cumulative product of noise schedules.

We adapt multimodal conditioning mechanism for medical abnormality synthesis by combining: (1) text descriptions $\mathcal{T}$ based on medical literature and anatomical terminology (e.g., ``pedunculated polyp with smooth surface'' or ``enlarged optic cup''), and (2) binary masks $M \in \{0, 1\}^{h \times w}$ derived from existing annotations indicating plausible abnormality locations.

The text descriptions are processed through a pre-trained text encoder $\tau$ to extract semantic embeddings:
\begin{equation}
e_{\text{text}} = \tau(\mathcal{T}) \in \mathbb{R}^{d_{\text{text}}}
\end{equation}

The reverse diffusion process learns to denoise while respecting both textual and spatial constraints:
\begin{equation}
p_\theta(z_{t-1} | z_t) = \mathcal{N}(z_{t-1}; \mu_\theta(z_t, t, e_{\text{text}}, M, z_n), \sigma_t^2 \mathbf{I})
\end{equation}

Our denoising network $\epsilon_\theta$ predicts the noise component:
\begin{equation}
\hat{\epsilon} = \epsilon_\theta(z_t, t, e_{\text{text}}, M, z_n)
\end{equation}

The training objective encourages abnormality generation within masked regions while preserving surrounding anatomy:
\begin{equation}
\resizebox{0.4\textwidth}{!}{$\mathcal{L}_{\text{gen}} = \mathbb{E}_{t,\epsilon,z_0} \left[| M \odot (\epsilon - \hat{\epsilon}) |^2 + \lambda_{\text{preserve}} | (1-M) \odot (z_t - \sqrt{\bar{\alpha}_t} z_n) |^2\right]$}
\end{equation}

During inference, we employ classifier-free guidance to strengthen conditioning:
\begin{equation}
\hat{\epsilon}_{\text{guided}} = \hat{\epsilon}_{\text{uncond}} + s \cdot (\hat{\epsilon}_{\text{cond}} - \hat{\epsilon}_{\text{uncond}})
\end{equation}

\subsection{Latent-Space Segmentation for Quality Validation}

To ensure generated abnormalities are accurately localized, we employ a segmentation network that operates directly in the latent space, validating the spatial consistency of generated content without computational overhead of repeated encoding/decoding.

Given a generated abnormal image $I_a$, we obtain its latent representation $z_a = \mathcal{E}(I_a)$ using the frozen VAE encoder. The segmentation network $\mathcal{S}_\phi: \mathbb{R}^{h \times w \times d} \to \mathbb{R}^{h \times w \times c}$ predicts segmentation masks directly in this compressed space.

We employ a dual-encoder strategy: alongside the frozen VAE encoder $\mathcal{E}$, we introduce a trainable vision encoder $\mathcal{V}_\psi$ that adapts to medical imaging characteristics:
\begin{equation} 
f_{\text{vision}} = \mathcal{V}_\psi(z_a) \in \mathbb{R}^{h \times w \times d'} 
\end{equation}

Following recent advances in diffusion-based segmentation, we train our network using a diffusion objective in latent space. During training, we add noise to the ground-truth segmentation latent to obtain $z_t$, and train the network to denoise:
\begin{equation}
z_t = \sqrt{\bar{\alpha}_t} z_0 + \sqrt{1-\bar{\alpha}_t} \epsilon
\end{equation}

The key insight is that for segmentation—unlike generation—we can directly estimate the clean latent in a single step. The network predicts the noise $\hat{\epsilon}_{\text{seg}}$, from which we derive:
\begin{equation} 
\hat{z}_0 = \frac{1}{\sqrt{\bar{\alpha}_t}}(z_t - \sqrt{1-\bar{\alpha}_t} \hat{\epsilon}_{\text{seg}}) 
\end{equation}

This enables single-step inference during validation, dramatically improving efficiency compared to iterative sampling. The training objective combines noise prediction with direct latent supervision:
\begin{equation} 
\mathcal{L}_{\text{seg}} = |\epsilon - \hat{\epsilon}_{\text{seg}}|_1 + \lambda_{\text{latent}} |z_0 - \hat{z}_0|_1 + \lambda_{\text{dice}} \mathcal{L}_{\text{Dice}}(\mathcal{D}(\hat{M}_{\text{latent}}), M_{\text{gt}}) 
\end{equation}
where $\hat{M}_{\text{latent}} = \mathcal{S}_\phi(f_{\text{fused}})$ is the predicted segmentation in latent space.

For feature fusion, we concatenate the latent representations from both encoders:
\begin{equation} 
f_{\text{fused}} = \psi([z_a; f_{\text{vision}}])
\end{equation}
where $\psi$ represents learned projection layers. This preserves spatial correspondence while avoiding the computational cost of cross-attention mechanisms, making the validation process efficient enough for real-time quality assessment during generation.

\subsection{Integrated Pipeline and Quality Control}

Our complete framework operates as an iterative generation-validation loop that ensures high-quality synthetic data production. We create text descriptions based on established medical terminology for each abnormality type. For colonoscopy, this includes variations like ``small sessile polyp,'' ``large pedunculated polyp with irregular surface,'' and ``flat lesion with unclear borders.'' For fundus imaging, descriptions span from ``early glaucomatous cupping'' to ``advanced rim loss with vessel displacement.''

For each normal image $I_n$ in our dataset, we generate multiple abnormal variants using different text prompts $\mathcal{T}_i$, masks $M_i$, and random seeds. The generation process follows the reverse diffusion described in Section 3.1, producing diverse synthetic examples through different noise initializations.

Each generated image $I_a^{(i,j)}$ is validated using our segmentation network to ensure accurate abnormality localization. We first encode the generated image to obtain $z_a^{(i,j)} = \mathcal{E}(I_a^{(i,j)})$, then compute the IoU between the intended mask and the predicted segmentation:
\begin{equation} 
\text{IoU}^{(i,j)} = \text{IoU}(M_i, \mathcal{D}(\hat{M}_{\text{latent}}^{(i,j)}))
\end{equation}
where $\hat{M}_{\text{latent}}^{(i,j)} = \mathcal{S}_\phi(f_{\text{fused}}^{(i,j)})$ is the predicted segmentation mask in latent space, and $f_{\text{fused}}^{(i,j)} = \psi([z_a^{(i,j)}; f_{\text{vision}}^{(i,j)}])$ combines features from both encoders.

Only generations exceeding a quality threshold $\theta_{\text{quality}} = 0.7$ are retained, ensuring spatial accuracy. 
This threshold was empirically determined through validation experiments where we evaluated thresholds from 0.5 to 0.9, finding that 0.7 optimally balanced generation diversity (65\% acceptance rate) with localization precision and downstream task performance. Lower thresholds risked anatomically implausible placements, while higher thresholds overly restricted data diversity. The validated synthetic pairs $(I_a^{(i,j)}, M_i)$ satisfying $\text{IoU}^{(i,j)} > \theta_{\text{quality}}$ are combined with real training data for model training, effectively augmenting the dataset with anatomically plausible abnormalities.

\section{Experimental Details}
\label{sec:experiments}

\subsection{Datasets}
We evaluate our framework on three medical imaging datasets that exemplify the class imbalance challenge in clinical imaging. For polyp detection, we use CVC-ClinicDB~\cite{bernal2015wm} containing 612 colonoscopy frames from 29 video sequences, and Kvasir-SEG~\cite{jha2019kvasir} with 1000 polyp-containing images selected from colonoscopy examinations. For glaucoma assessment, we employ REFUGE2~\cite{orlando2020refuge} containing 1200 fundus images with pixel-level optic disc and cup annotations.

\begin{table}[h!]
\caption{Dataset statistics showing class distribution and data splits}
\centering
\resizebox{0.48\textwidth}{!}{%
\begin{tabular}{l|l|l|l|l}
\multirow{2}{*}{\textbf{Dataset}} & \multirow{2}{*}{\textbf{Target}} & \multicolumn{2}{c|}{\textbf{Training}} & \multirow{2}{*}{\textbf{Test}} \\ \cline{3-4}
                                 &                                 & \textbf{Total} & \textbf{w/ Abnorm.} &  \\ \hline
CVC-ClinicDB~\cite{bernal2015wm} & Polyp                           & 488          & 488 (100\%)        & 124         \\ 
Kvasir-SEG~\cite{jha2019kvasir} & Polyp                           & 800          & 800 (100\%)        & 200         \\ 
REFUGE2~\cite{orlando2020refuge} & Optic Cup/Disc                 & 800          & 800 (100\%)        & 400         \\ 
\end{tabular}%
}
\end{table}

For synthetic data generation, we extract normal regions from these datasets by excluding annotated abnormalities, providing the healthy tissue backgrounds needed for controlled abnormality synthesis. The train/test splits follow the original dataset protocols to ensure fair comparison with existing methods.

\subsection{Evaluation Metrics}
We evaluate both the quality of generated synthetic data and the performance improvement in downstream segmentation tasks.

\textbf{Segmentation Performance.} We report standard medical image segmentation metrics:
\begin{itemize}
    \item Dice Similarity Coefficient (DSC): $\frac{2|P \cap G|}{|P| + |G|}$
    \item Intersection over Union (IoU): $\frac{|P \cap G|}{|P \cup G|}$
\end{itemize}
where $P$ and $G$ denote predicted and ground truth masks respectively. Since reducing missed diagnoses is critical in screening applications, we also report sensitivity and false negative rate (FNR = 1 - sensitivity) as clinically relevant metrics.

\textbf{Synthetic Data Quality.} To validate our generated abnormalities, we employ:
\begin{itemize}
    \item \textbf{Spatial accuracy}: IoU between intended generation masks and segmentation network predictions (as described in Section 3.3)
    \item \textbf{Visual realism}: LPIPS distance between generated and real abnormal images
    \item \textbf{Diversity}: Number of unique abnormality variations generated per normal image
\end{itemize}

\subsection{Implementation Details}
Our framework is implemented in PyTorch using the Diffusers library. Experiments are conducted on NVIDIA RTX 4090 GPUs (24GB memory). All images are resized to $256 \times 256$ pixels to match the VAE encoder requirements, which compresses them to $32 \times 32 \times 4$ latent representations with a downsampling factor of $f = 8$.

\textbf{Generation Module.} We employ Stable Diffusion XL (SDXL)~\cite{podell2023sdxl} using the SDXL-inpainting checkpoint optimized for masked generation. The diffusion model operates in the compressed latent space, fine-tuned for 50,000 steps with AdamW optimizer (learning rate $1 \times 10^{-5}$, batch size 8).

\textbf{Text Prompts.} We construct domain-specific prompt banks based on medical terminology:
\begin{itemize}
    \item \textbf{Polyps:} Morphological variations (``sessile polyp 5-10mm,'' ``pedunculated polyp with long stalk''), surface characteristics (``smooth surface,'' ``irregular lobulated surface''), and vascular patterns (``visible capillary pattern'')
    \item \textbf{Glaucoma:} Clinical features (``enlarged cup-to-disc ratio 0.7,'' ``inferior rim thinning,'' ``RNFL defect'')
    \item \textbf{Negative prompts:} Normal tissue descriptions (``healthy mucosa,'' ``normal optic disc'')
\end{itemize}

\textbf{Segmentation Network.} Built upon the Stable Diffusion U-Net architecture, the segmentation network processes $32 \times 32 \times 4$ latent representations. We initialize the encoder from pre-trained weights while training the decoder from scratch. Feature fusion employs $1 \times 1$ convolutions for concatenation. Training uses AdamW optimizer ($\beta_1 = 0.9$, $\beta_2 = 0.999$, weight decay $0.01$) with cosine learning rate schedule from $1 \times 10^{-5}$ to $1 \times 10^{-6}$. We use batch size 4 with gradient accumulation over 4 steps (effective batch size 16) for 100,000 iterations.

\textbf{Quality Control.} Generated images are accepted only if the segmentation network's predicted mask achieves IoU $> 0.7$ with the intended generation mask, ensuring spatial accuracy.
\begin{table*}[t!]
\centering
\caption{Performance comparison on medical segmentation benchmarks. Best results in bold, second best underlined.}
\label{tab:performance_comparison}
\begin{tabular}{@{}llcccc@{}}
\toprule
\textbf{Method Type} & \textbf{Method} & \textbf{CVC-ClinicDB}~\cite{bernal2015wm} & \textbf{Kvasir-SEG}~\cite{jha2019kvasir} & \textbf{REFUGE2}~\cite{orlando2020refuge} \\
                     &                 & Dice (\%) / IoU (\%) & Dice (\%) / IoU (\%) & Dice (\%) / IoU (\%) \\
\midrule
\multirow{3}{*}{\parbox{3cm}{Traditional\\Augmentation}} & U-Net~\cite{ronneberger2015u} & 88.3$\pm$1.2 / 81.4$\pm$1.8 & 86.7$\pm$1.5 / 79.2$\pm$2.1 & 80.1$\pm$1.7 / 72.3$\pm$2.2 \\
& U-Net (with Resnet50) & 90.1$\pm$0.9 / 83.7$\pm$1.4 & 88.4$\pm$1.2 / 81.5$\pm$1.6 & 82.4$\pm$1.3 / 74.8$\pm$1.8 \\
& TransUNet~\cite{chen2021transunet} & 91.8$\pm$0.7 / 85.9$\pm$1.1 & 90.2$\pm$0.9 / 83.8$\pm$1.3 & 85.6$\pm$1.0 / 78.2$\pm$1.4 \\
\midrule
\multirow{5}{*}{\parbox{3cm}{Diffusion-based\\Methods}} & MedSegDiff-V1~\cite{wu2024medsegdiff} & 93.2$\pm$0.6 / 88.1$\pm$0.9 & 92.3$\pm$0.7 / 86.4$\pm$1.1 & 86.3$\pm$0.8 / 78.2$\pm$1.1 \\
& MedSegDiff-V2~\cite{wu2025medsegdiff} & 93.8$\pm$0.5 / 88.9$\pm$0.8 & 92.8$\pm$0.6 / 87.1$\pm$0.9 & 85.9$\pm$0.7 / 79.6$\pm$1.0 \\
& Diff-Trans~\cite{chowdary2023diffusion} & 95.4$\pm$0.3 / 92.0$\pm$0.4 & 94.6$\pm$0.4 / 91.6$\pm$0.5 & 88.7$\pm$0.6 / 81.5$\pm$0.7 \\
& SDSeg~\cite{lin2024stable} & \underline{95.8$\pm$0.2} / \underline{92.6$\pm$0.3} & \underline{94.9$\pm$0.3} / \underline{92.1$\pm$0.4} & \underline{89.4$\pm$0.4} / \underline{81.8$\pm$0.6} \\
& \methname\textbf{(Ours)} & \textbf{96.4$\pm$0.3} / \textbf{93.2$\pm$0.4} & \textbf{95.6$\pm$0.3} / \textbf{92.8$\pm$0.4} & \textbf{90.2$\pm$0.4} / \textbf{82.6$\pm$0.5} \\
\bottomrule
\end{tabular}
\end{table*}

\subsection{Inference Pipeline}
During inference, we generate synthetic abnormalities through a two-stage process:

\textbf{Generation Stage.} For each normal image, we perform conditional inpainting using SDXL with the following settings:
\begin{itemize}
    \item 50 denoising steps for generation quality
    \item Classifier-free guidance scale of 7.5 (balancing fidelity and diversity)
    \item 5-10 variants per image using different text-mask combinations
\end{itemize}

\textbf{Validation Stage.} Each generated image undergoes quality validation through our segmentation network:
\begin{itemize}
    \item Single-step inference using the trained latent estimation
    \item IoU computation between predicted and intended masks
    \item Acceptance threshold: IoU $> 0.7$
\end{itemize}

Images failing validation are discarded rather than regenerated, as our experiments show consistent generation quality across random seeds. The complete pipeline achieves throughput of approximately 1000 validated images per hour on a single RTX 4090 GPU.
\section{Main Results}
\label{sec:results}

\subsection{Comparison with State-of-the-Art Methods}

Table~\ref{tab:performance_comparison} presents comprehensive evaluation results across three medical segmentation benchmarks. Our method achieves state-of-the-art performance with Dice coefficients of 96.4\% on CVC-ClinicDB, 95.6\% on Kvasir-SEG, and 90.2\% on REFUGE2, consistently outperforming all baseline approaches.

Compared to traditional augmentation approaches, our method shows substantial improvements. The baseline U-Net achieves only 88.3\% Dice on CVC-ClinicDB, while our approach reaches 96.4\% an 8.1\% absolute improvement. Similar gains are observed across all datasets: 8.9\% improvement on Kvasir-SEG and 10.1\% on REFUGE2. Even against stronger baselines like TransUNet (91.8\% on CVC-ClinicDB), our method maintains a significant 4.6\% advantage.

Among recent diffusion-based methods, our approach sets new benchmarks. We surpass SDSeg by 0.6\% on CVC-ClinicDB (96.4\% vs. 95.8\%), 0.7\% on Kvasir-SEG (95.6\% vs. 94.9\%), and 0.8\% on REFUGE2 (90.2\% vs. 89.4\%). While these improvements over SDSeg appear incremental, they are statistically significant ($p<0.05$). More substantial gains are achieved over MedSegDiff-V2, with improvements of 2.6\%, 2.8\%, and 4.3\% across the three datasets respectively.

The IoU metrics further validate our superior performance, achieving 93.2\%, 92.8\%, and 82.6\% on the three benchmarks—consistently 0.6-3.0\% higher than the next best method. Notably, our approach exhibits the lowest standard deviations (±0.3-0.4\%) among all methods, indicating robust and stable predictions essential for clinical deployment. This stability represents a marked improvement over traditional augmentation approaches, which show variations of ±0.7-2.2\%.

\subsection{Clinical Performance Analysis}

Table~\ref{tab:lesion_level_detection} presents lesion-level detection performance, focusing on clinically challenging cases where reducing false negatives is critical for patient outcomes. In screening applications, false negative rates (FNR) directly correlate with missed diagnoses, making even modest improvements clinically relevant.

For small polyps (<5mm), our method reduces FNR to 23.2\% compared to 25.9\% for SDSeg and 31.7\% for baseline U-Net. While the 2.7 percentage point improvement over SDSeg appears modest, this translates to detecting approximately 11\% more small polyps that would otherwise be missed (2.7/25.9). Given that small polyps can progress to colorectal cancer, this reduction in missed cases has potential clinical significance.

Flat lesions show similar improvements, with FNR decreasing from 31.1\% (SDSeg) to 27.6\% (ours). The 3.5 percentage point reduction means approximately 11\% fewer flat lesions are missed compared to SDSeg. These morphologically subtle abnormalities benefit from our synthetic training examples generated with descriptors like "subtle mucosal elevation" and "irregular surface pattern."

For early glaucoma detection (CDR 0.5-0.6), FNR improves from 22.7\% (SDSeg) to 20.7\% (ours). While this 2.0 percentage point improvement is the smallest among the three categories, early glaucoma detection is particularly challenging as these cases sit at the borderline of normal variation.

\begin{table*}[h!]
\centering
\caption{Lesion-level detection performance for challenging abnormality types.}
\label{tab:lesion_level_detection}
\begin{tabular}{@{}llcccc@{}}
\toprule
\textbf{Abnormality Type} & \textbf{Method} & \textbf{Sensitivity (\%)} & \textbf{FNR (\%)} & \textbf{Relative FNR} & \textbf{Specificity (\%)} \\
& & & & \textbf{Reduction (\%)} & \\
\midrule
\multirow{3}{*}{Small Polyps ($<5mm$)} 
& U-Net & 68.3 ± 3.2 & 31.7 & --- & 94.2 ± 1.8 \\
& SDSeg & 74.1 ± 2.4 & 25.9 & 18.3 & 95.6 ± 1.4 \\
& \textbf{\methname} & \textbf{76.8 ± 2.1} & \textbf{23.2} & \textbf{26.8} & \textbf{96.1 ± 1.2} \\
\midrule
\multirow{3}{*}{Flat Lesions}
& U-Net & 61.5 ± 3.8 & 38.5 & --- & 92.8 ± 2.1 \\
& SDSeg & 68.9 ± 3.1 & 31.1 & 19.2 & 94.3 ± 1.7 \\
& \textbf{\methname} & \textbf{72.4 ± 2.8} & \textbf{27.6} & \textbf{28.3} & \textbf{94.9 ± 1.5} \\
\midrule
\multirow{3}{*}{Early Glaucoma (CDR 0.5-0.6)}
& U-Net & 72.1 ± 2.9 & 27.9 & --- & 88.4 ± 2.3 \\
& SDSeg & 77.3 ± 2.3 & 22.7 & 18.6 & 90.2 ± 1.9 \\
& \textbf{\methname} & \textbf{79.3 ± 2.0} & \textbf{20.7} & \textbf{25.8} & \textbf{91.1 ± 1.6} \\
\bottomrule
\end{tabular}
\end{table*}

\subsection{Computational Efficiency}

Table~\ref{tab:computational_efficiency} presents computational requirements across different methods. Since our framework consists of two components—synthetic data generation and segmentation—we report metrics for both, while comparison methods only perform segmentation.

For segmentation (the primary task), our network achieves 12.5 samples/second inference speed using single-step latent estimation. This is 40× faster than MedSegDiff (0.30-0.31 samples/s), which requires iterative denoising, and 1.5× faster than SDSeg (8.36 samples/s), while also achieving superior accuracy as shown in Table~\ref{tab:performance_comparison}.

Our generation component, used offline to create synthetic training data, processes 0.40 samples/second. While slower than segmentation, this is a one-time preprocessing step that doesn't affect deployment speed. Once synthetic data is generated and validated, only the fast segmentation network is needed for clinical use.

Training efficiency represents another key advantage. Both our components require only single-GPU training (8 hours for segmentation, 15 hours for generation), compared to 4-GPU setups needed by MedSegDiff (48-49 hours) and Diff-UNet (16 hours).

\begin{table}[h!]
\centering
\caption{Computational efficiency comparison. Note: Other methods perform only segmentation, while ours includes both generation (offline) and segmentation components.}
\label{tab:computational_efficiency}
\resizebox{0.45\textwidth}{!}{%
\begin{tabular}{@{}lccc@{}}
\toprule
\textbf{Method} & \textbf{Training Time} & \textbf{GPU} & \textbf{Inference Speed} \\
& \textbf{(hours)} & \textbf{Requirement} & \textbf{(samples/s)} \\
\midrule
MedSegDiff-V1 & ~48 & 4× GPU & 0.30 \\
MedSegDiff-V2 & ~49 & 4× GPU & 0.31 \\
Diff-UNet & ~16 & 4× GPU & 0.87 \\
SDSeg & ~12 & 1× GPU & 8.36 \\
\midrule
\textbf{\methname(Segmentation)} & ~8 & 1× GPU & \textbf{12.50} \\
\textbf{\methname(Generation)}\textsuperscript{*} & ~15 & 1× GPU & 0.40 \\
\bottomrule
\end{tabular}
}
\tablefootnote{*Generation is an offline preprocessing step for creating synthetic training data.}
\end{table}

\begin{table}[htbp]
\centering
\caption{Automated quality metrics for generated images. Lower is better for FID and LPIPS.}
\label{tab:automated_metrics}
\resizebox{0.4\textwidth}{!}{%
\begin{tabular}{@{}lcc@{}}
\toprule
\textbf{Metric} & \textbf{Accepted} & \textbf{Rejected} \\
\midrule
FID Score $\downarrow$ & $32.4 \pm 2.8$ & $78.6 \pm 6.3$ \\
LPIPS $\downarrow$ & $0.245 \pm 0.03$ & $0.512 \pm 0.08$ \\
Validation IoU & $0.76 \pm 0.11$ & $0.38 \pm 0.19$ \\
Acceptance Rate & $78.3\%$ & $21.7\%$ \\
\bottomrule
\end{tabular}
}
\end{table}


\subsection{Quality Assessment of Generated Abnormalities}

\begin{figure}[t]
\centering
\includegraphics[width=\linewidth]{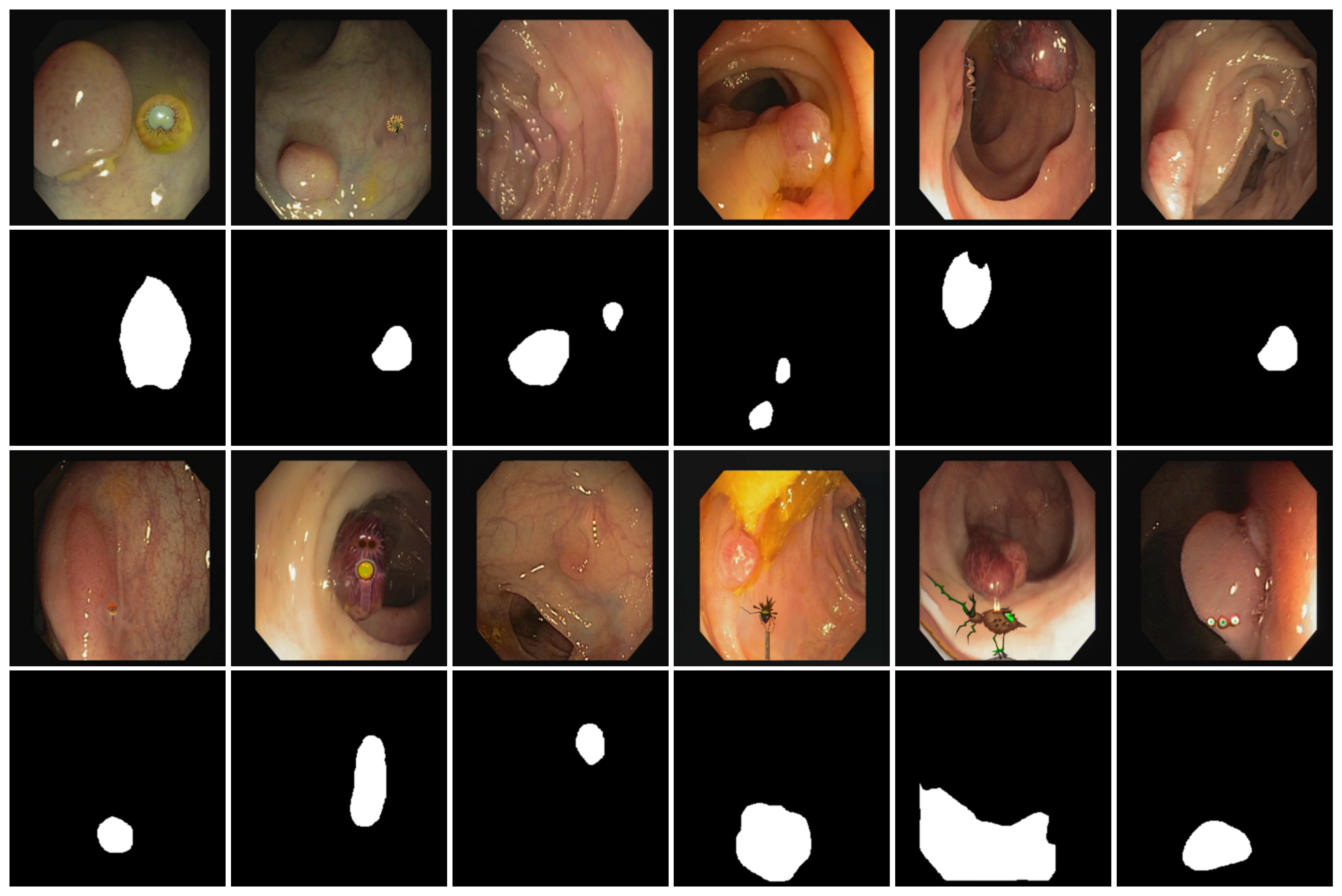}
\caption{Examples of synthetic abnormalities generated by our framework. Shows various polyps (CVC-ClinicDB dataset) created using different text prompts such as "sessile polyp," "pedunculated polyp." Each image represents a different generated abnormality with varying characteristics.}
\label{fig:augmented}
\end{figure}

We evaluate the quality of generated abnormalities through automated metrics to ensure they are suitable for training augmentation. Figure~\ref{fig:augmented} shows example synthetic abnormalities generated by our framework across different types and datasets. Tables~\ref{tab:automated_metrics} presents quantitative quality evaluations.

Automated metrics demonstrate that accepted samples meet our quality threshold for training augmentation. FID scores of $32.4$ compared to real abnormal images, though these metrics were designed for natural images rather than medical data, with LPIPS values of $0.245$ indicating perceptual distance from real images. The large gap between accepted and rejected samples validates our quality control mechanism.

The validation IoU between intended and detected abnormalities averages $0.76$ for accepted samples, confirming spatial control. Rejected samples show an IoU of $0.38$, justifying our $0.7$ threshold. Approximately $20$–$25\%$ of initial generations fail validation, primarily for challenging cases like flat polyps where subtle features are difficult to synthesize.

\section{Ablation Study}
\label{sec:ablationstudy}
\subsection{Component Analysis}
We conduct ablation studies to validate the contribution of each framework component. Table~\ref{tab:ablation_study} presents results on CVC-ClinicDB, examining the impact of removing key elements. Removing text guidance reduces Dice by 1.1\%, showing the benefit of controlled generation. Without the validation loop, performance drops by 1.4\% Dice due to the inclusion of lower-quality synthetic samples. Using only basic prompts instead of diverse descriptions decreases performance by 0.8\%, indicating that prompt variety helps capture different abnormality presentations.

The latent estimation mechanism is essential for efficiency. Without it, inference speed drops from 12.5 to 0.31 samples/s due to iterative denoising, making the validation process impractical. Most significantly, training without synthetic augmentation results in an 8.1\% Dice decrease and increases FNR from 23.2\% to 31.7\%, demonstrating the clear value of our data generation approach for addressing class imbalance.
\begin{table}[htbp]
\centering
\caption{Ablation study results on CVC-ClinicDB dataset.}
\label{tab:ablation_study}
\resizebox{\columnwidth}{!}{%
\begin{tabular}{@{}lcccc@{}}
\toprule
\textbf{Configuration} & \textbf{Dice (\%)} & \textbf{IoU (\%)} & \textbf{FNR (\%)} & \textbf{Inference Speed} \\
& & & & (samples/s) \\
\midrule
\textbf{\methname (Full)} & \textbf{96.4 ± 0.3} & \textbf{93.2 ± 0.4} & \textbf{23.2} & \textbf{12.5} \\
\midrule
w/o Text Guidance & 95.3 ± 0.4 & 91.8 ± 0.5 & 25.1 & 12.5 \\
w/o Validation Loop & 95.0 ± 0.5 & 91.4 ± 0.6 & 25.8 & N/A \\
w/o Diverse Prompts & 95.6 ± 0.4 & 92.3 ± 0.5 & 24.3 & 12.5 \\
w/o Latent Estimation & 96.1 ± 0.3 & 92.8 ± 0.4 & 23.7 & 0.31 \\
w/o Synthetic Data & 88.3 ± 1.2 & 81.4 ± 1.8 & 31.7 & 12.5 \\
\bottomrule
\end{tabular}%
}
\end{table}

\subsection{Impact of Synthetic Data Volume}

To further understand the contribution of synthetic data, we analyze performance as a function of the number of generated images added to training. Table~\ref{tab:augmentation_volume} shows results with varying amounts of synthetic augmentation.

\begin{table}[h!]
\centering
\caption{Performance with varying amounts of synthetic augmentation on CVC-ClinicDB.}
\label{tab:augmentation_volume}
\resizebox{0.48\textwidth}{!}{%
\begin{tabular}{@{}lcccc@{}}
\toprule
\textbf{Training Configuration} & \textbf{Real} & \textbf{Synthetic} & \textbf{Dice (\%)} & \textbf{IoU (\%)} \\
& \textbf{Images} & \textbf{Images} & & \\
\midrule
Baseline (no augmentation) & 488 & 0 & 88.3±1.2 & 81.4±1.8 \\
Standard augmentation only\textsuperscript{*} & 488 & 0 & 90.1±1.0 & 83.5±1.5 \\
\midrule
\methname + 0.5× synthetic & 488 & 244 & 92.8±0.7 & 87.3±1.1 \\
\methname + 1× synthetic & 488 & 488 & 94.6±0.5 & 90.1±0.8 \\
\methname + 2× synthetic & 488 & 976 & 95.7±0.4 & 92.0±0.6 \\
\textbf{\methname + 3× synthetic} & \textbf{488} & \textbf{1464} & \textbf{96.4±0.3} & \textbf{93.2±0.4} \\
\methname + 5× synthetic & 488 & 2440 & 96.3±0.3 & 93.1±0.4 \\
\bottomrule
\end{tabular}
}
\tablefootnote{*Standard augmentation includes rotation, flipping, scaling, and elastic deformation.}
\end{table}


Performance improves steadily as synthetic images are added, with the most substantial gains occurring up to 3× the original dataset size (1,464 synthetic images). Beyond this point, performance plateaus, suggesting we have achieved sufficient diversity for this dataset. Even with just 0.5× synthetic data (244 images), we observe a 2.7\% Dice improvement over standard augmentation, demonstrating the quality of our generated samples.

The diminishing returns beyond 3× augmentation indicate that our generation process effectively captures the variety of abnormality presentations. This finding has practical implications: generating 3× synthetic data provides optimal performance while minimizing computational costs. These results confirm that our text-guided generation with quality validation provides meaningful improvements through high-quality synthetic samples rather than simply increasing data volume.

\section{Conclusion}
\label{sec:conclusion}


We presented a framework that addresses data scarcity in medical image segmentation by combining text-guided diffusion-based generation with automatic quality validation. The framework reduces false negative rates for challenging cases like small polyps and flat lesions by up to 28\%, while maintaining efficient inference at 12.5 samples/second through single-step latent estimation. Our ablation studies confirm that text guidance, quality validation, and 3× synthetic augmentation each contribute meaningfully to performance gains. By requiring only consumer grade GPUs and generating high-quality synthetic training data without manual annotation, our method provides a practical solution for addressing class imbalance in medical imaging applications.

\section*{Acknowledgements}

This study was carried out within the PNRR research activities of the
consortium iNEST (Interconnected North-Est Innovation Ecosystem) funded by the European Union Next-GenerationEU (Piano Nazionale di Ripresa e Resilienza (PNRR) – Missione 4 Componente 2, Investimento 1.5 – D.D. 1058  23/06/2022, ECS\_00000043).